\documentclass[]{elsarticle}

\usepackage{lineno}
\usepackage{hyperref}
\usepackage{amsmath}
\usepackage{graphicx,psfrag,epsf}
\usepackage{enumerate}
\usepackage{natbib}

\usepackage{fix-cm}
\usepackage{framed}
\usepackage{multirow}
\usepackage{rotating}
\usepackage{tabularx}
\usepackage{float}
\usepackage{ctable}
\usepackage{amsfonts}
\usepackage{arydshln}
\usepackage{enumitem}
\usepackage{lipsum}
\usepackage{setspace}

\newcommand{\ex}{{\mathbb E}}

\newcommand*\vtick{\textsc{\char13}}

\makeatletter
\def\ps@pprintTitle{%
	\let\@oddhead\@empty
	\let\@evenhead\@empty
	\def\@oddfoot{}%
	\let\@evenfoot\@oddfoot}
\makeatother

\begin{document}

\begin{frontmatter}

\title{Temporal Clustering of Time Series via Threshold Autoregressive Models: Application to Commodity Prices}

\author[mymainaddress,mysecondaryaddress]{Sipan Aslan\corref{mycorrespondingauthor}}
\cortext[mycorrespondingauthor]{Corresponding author}
\ead{sipan@metu.edu.tr}

\author[mymainaddress]{Ceylan Yozgatligil}
\author[mymain2address]{Cem Iyigun}

\address[mymainaddress]{Department of Statistics, Middle East Technical University}
\address[mysecondaryaddress]{Faculty of Sciences, Yuzuncu Yil University}
\address[mymain2address]{Department of Industrial Engineering, Middle East Technical University}
\begin{abstract}
This study aimed to find temporal clusters for several commodity prices using the threshold non-linear autoregressive model. It is expected that the process of determining the commodity groups that are time-dependent will advance the current knowledge about the dynamics of co-moving and coherent prices, and can serve as a basis for multivariate time series analyses. The clustering of commodity prices was examined using the proposed clustering approach based on time series models to incorporate the time varying properties of price series into the clustering scheme. Accordingly, the primary aim in this study was grouping time series according to the similarity between their Data Generating Mechanisms (DGMs) rather than comparing pattern similarities in the time series traces. The approximation to the DGM of each series was accomplished using threshold autoregressive models, which are recognized for their ability to represent nonlinear features in time series, such as abrupt changes, time-irreversibility and regime-shifting behavior. Through the use of the proposed approach, one can determine and monitor the set of co-moving time series variables across the time dimension. Furthermore, generating a time varying commodity price index and sub-indexes can become possible. Consequently, we conducted a simulation study to assess the effectiveness of the proposed clustering approach and the results are presented for both the simulated and real data sets. 
\end{abstract}


\end{frontmatter}


\section{Introduction}
\label{intro}
The movement of commodity prices and the associated dynamics are interrelated with economics and directly affect many industries. For example, energy based commodities constitute the main input cost for firms and households; changes in agricultural food prices affect the purchasing power parity; and base metals such as copper and aluminum serve as primary raw materials for industrial production and the building trade. Inflation and growth rate dynamics are linked to changes in commodity prices and the form of the link is time varying. An avenue of research which is receiving a lot of attention both from monetary and fiscal policy makers and from academia is that into the dynamics and the endogeneity of commodity prices to macroeconomic variables and monetary developments, such as expected growth, expected inflation, interest rates, and currency movements  \citep[see, among the others,][]{furlong96, pindyck2004volatility, frank06, chen2014drives, chenzivot14inflation}. Moreover, understanding commodity price behavior has become more important for commodity market participants since the financialization of commodities \citep{tang12,cheng2014financialization}. Findings suggest that determining commodity groups that are co-moving with changing dynamics will find an application in inflation prediction and investors' portfolios \citep{tang12,sensoy2015dynamic}. Hence, determining price groups that are time-dependent can deepen the knowledge about the dynamics of co-moving and coherent prices, and it can serve as a basis for other statistical analyses, such as causality, co-integration and multivariate analyses, from which the underlying causes of co-movements can be investigated. Generating commodity price indices (e.g. RICI, TRCCI, S$\&$P GSCI) that rely on time-dependently determined price groups are also possible by assigning weights for each commodity or for commodity groups.

Research on co-movements of commodity prices has been the subject of interest following the seminal work of Pindyck and Rotemberg \citep{pind90}, in which they concluded that seemingly unrelated commodity prices tend to move together. They hypothesized that this price behavior was an "excess" co-movement that was not related to macroeconomic fundamentals, such as inflation, industrial production, interest rates or exchange rates, but rather due to herding behavior, where traders speculate on commodities for no plausible economic reason \citep{pind90}. Since then, a number of researches have been focused on the co-movement of commodity prices, in which consideration have been given to various aspects of the price movements via econometric model specifications, such as the effects of macroeconomic variables, structural changes and the volatility of prices \citep[see, among the others,][]{ai2006comovement,lescar09,byrne2013primary,stee13, poncela2014common,matesanz2014co,fer15,ross15}. For example, Ai et al. \citep{ai2006comovement} presented evidence against the herding behavior of commodities: the data suggested that the majority of the co-movements were related with commodity factors such as supply and demand. They claimed that information about price movements was insufficient to understand commodity markets or develop a commodity price model. Similarly, according to the findings of Lescaroux \citep{lescar09}, for oil and six particular metals, price movements did not support the hypothesis of "excess" co-movement, but rather demonstrated that the tendency of commodity prices to co-move was related to the tendency of their fundamental factors to move together.

Nevertheless, Byrne et al. \citep{byrne2013primary} reported significant evidence in favor of co-movement in commodity prices and identified a common factor utilizing the non-stationary Panel factor (PANIC) model from Bai and Ng \citep{bai2004panic} and Factor Augmented Vector Autoregression (FAVAR) framework from Bernanke et al. \citep{benber2005favar}. Examples of a sector and intra-sector based assessment of price co-movements for energy, non-energy and metal type commodities can be found in \citep{ross15,fer15,sensoy2015dynamic}. Rossen \citep{ross15} focused and analyzed the prices of different groups of metals, namely non-ferrous metals, light metals and precious metals,  using common statistical methods: concordance statistics, distance correlation coefficients, cross-correlations and co-integration analyses. The findings suggested that co-movement is not viable among metal prices, but it can be validated within specific groups of metals. Fernandez \citep{fer15} focused on a new measure of global co-movement by determining the average partial autocorrelation of one commodity with others. Several commodity categories and their influence on other commodity markets were analyzed and the results showed that there has been a strong co-movement among the nominal returns of metals since 2003 and that the co-movement among unrelated commodity returns has been negligible except for the period of global financial crisis in 2007-2010. Similarly, Sensoy et al. \citep{sensoy2015dynamic} evidenced a dynamic convergence of commodity futures returns for three groups of commodities (precious and industrial metals, and energy) by utilizing a dynamic equicorrelation (DECO)-generalized autoregressive conditional heteroscedasticity (GARCH) model. The example for co-movement analyses in a wide group of commodity prices via clustering perspective can be found in \citep{matesanz2014co}. In their paper, Matesanz et al. \citep{matesanz2014co} implemented the hierarchical clustering based on a model free dissimilarity measure and results provided for describing the dynamics of co-movements and temporal interdependencies of commodities. Moreover, throughout the majority of literature about the co-movement of commodity prices, attention was drawn to the temporal variability of price behavior.

This study aimed to address the question of determining the time varying commodity price groups that may have a co-movement property and do so in a way that differed from existing literature in two ways. First, we proposed the time series clustering approach for grouping of commodity prices in which we would utilize the time series model-based and goal-oriented feature vectors that are able to represent the time varying behavior of commodity prices. In addition, since the clustering methodology does not assume, if it is not designated, any category based evaluation, it is possible to form a group in which its members originate from different commodity categories. Thus, seemingly unrelated commodities, such as gold and natural gas, may fall into the same cluster. Second, the temporal variability of prices was examined by a regime switching concept and we proposed the use of a threshold autoregressive (TAR) time series model to incorporate the time varying properties of price series into the clustering task. More specifically, each time series was to be formed into a vector that represents the point in the generated high dimensional space and then the vectors on this space would be clustered/grouped. It should also be noted that to represent the temporal dynamics of time series, and therefore, clusters, vectors that represent each time series could be re-generated at each time point. 

This paper is organized as follows: the motivation behind the study is explained in section 2, the methodology and proposed approach are given in section 3 and the real data used in the study are presented in section 4, which is mainly devoted to the results of the simulation and the application study. The conclusion and future work are summarized in section 5.

\section{Motivation and Contribution of the Study}
\label{sec:2}
Time series clustering is the unsupervised grouping of a set of unlabeled time series into homogeneous clusters in such a way that the within-group dissimilarity is minimized and the between-group dissimilarity is maximized at the end. Various approaches and procedures have been developed and used to cluster time series from different fields, such as economics, finance, bioinformatics, neuroscience, and climatology; however, most of them share a common perspective, in which they use or modify the algorithms for the clustering of static data by converting time series data into the form of static data. In this process, feature extraction from time series can be implemented and it is the extracted information (i.e. static information such as mean and variance) that is then used in clustering instead of the raw data. Although the intention is for the extracted features to be representative of the time series, information that has static characteristics (i.e. time independent) is not sufficient to represent the time series when their temporal behavior is considered. Therefore, besides the technique used in clustering, the performance of the clustering approach of time series is highly dependent on feature selection. The way of feature extraction is a continuing issue that needs to be addressed carefully. Comprehensive surveys based on different aspects of time series clustering can be seen in \citep{liao2005clustering, fu2011review, aghabozorgi2015time}.

The perfect  clustering of multiple time series, which also implies classification in the idealized case, can be achieved if the underlying data generating mechanisms (DGMs) are thoroughly known. Since we observe limited realizations of the underlying processes, determining the true model (i.e. the actual DGM) is usually not feasible in real cases. However, statistical inferences can be accessible after obtaining good approximations to those DGMs. In this study, the proposed time series clustering approach aimed to incorporate the time-dependent information of the time series that was derived from approximations to their true DGMs. Here, the approximation to DGMs was done using linear and non-linear time series models and the associations of time series with these time series models were used in the proposed time series clustering approach. Thus, the feature extraction phase or the feature vector formation of the proposed clustering approach mainly relied on time series models. Two main objectives were expected to be achieved from the formation of feature vectors: firstly, to ensure the time-dependent characteristics were well represented at feature vectors in a proper manner; and secondly, to provide comparable and distinctive inputs to feature vectors from DGM approximations. In this study, one of the approximations to the DGM of each time series was accomplished by the TAR models detailed in Section 3.1. They are known for their dynamic structure and ability to represent nonlinear features in time series, such as abrupt changes, asymmetry and especially, regime shifting behavior (time varying state phenomena).

Another motivation of this study was to gain an insight into co-moving commodities by investigating temporal price groups in commodities through a time series clustering perspective. Since the aim of the proposed clustering approach was to group time series according to their time-dependent behavioral similarities, the group members in the same cluster that share similar dynamics (i.e. DGM) would provide useful information for further investigation of co-movement evaluations. In the literature, to the best of authors' knowledge, a time series model based clustering framework has not been reported for commodity prices. 

\section{Methodology}
\label{sec:3}
The primary concern for this paper was grouping time series with respect to the similarity/dissimilarity between their DGM approximations instead of considering their commonly shared patterns (i.e. time series traces, static information). Since identical non-deterministic DGMs can produce different patterns (i.e. traces) as a result of their stochastic nature, finding a coherent time series group with respect to a trace-like pattern similarity/dissimilarity can lead to inappropriate conclusions. In order to distinguish time series with respect to their actual DGM approximations, we need a rich environment (i.e. a multifaceted time series model) to provide distinctive outcomes for clustering. Thus, instead of aiming to find a true DGM, the associations of time series with this environment can be observed and used for clustering. Fortunately, by having a switching mechanism that allows time varying states, accounting nonlinearities with a relatively simple structure, and enabling an easy implementation procedure, TAR models can provide comparable outputs for each time series. 

In brief, to determine the clusters of time series, non-linear TAR model outputs, Autoregressive (AR) model outputs, the sample autocorrelation function (acf), the partial autocorrelation function (pacf) and the cross-correlation function (ccf) of each time series were combined within the proposed clustering approach. The TAR model and the proposed approach are explained in the following sub-sections. 
\subsection{Threshold Autoregressive Model (TAR)}
\label{sec:3.1}
The TAR model, introduced by Tong and Lim \citep{tong1980threshold}, is motivated  by several nonlinear characteristics commonly observed in practice such as asymmetry in declining and rising patterns of a time-dependent process. The model aims to determine the time varying behavior of a time series process by switching regimes (states) via a threshold variable. Thus, unlike the linear and autoregressive time reversible time series models, the TAR modeling perspective seems to have a satisfactory way of analyzing time irreversible and complex systems due to its handling of the complexity within different but simpler linear subsystems that are connected by a threshold process. The capability of TAR models to generate and capture non-linear dynamics, limit cycles, severe jumps and asymmetries is exemplified in diverse fields and comprehensively discussed in Tong \citep{tong1990non}. The subsequent effect and reflection of the threshold non-linearity concept and TAR modelling in the fields of economics and finance are reviewed in Hansen \citep{hansen2011threshold} and Chen \citep{chen2011review}.

A general TAR($k$) model where $k$ denotes the number of regimes can be represented as
\begin{equation}\label{eq1}
	y_{t}  =\sum_{j=1}^{k} \left [ (\phi_{0}^{(j)} + \sum_{i=1}^{p_j}\phi_{i}^{(j)}y_{t-i} \ + \ \varepsilon_{t}^{(j)}) \ I(r_{j-1}<z_{t-d}\leq r_{j}) \right]
\end{equation}
where  $y_{t}$ represents the time series process; $j=1,\ldots,k$ and $k$ is the 
number of regimes; $p_j$ is for $j^{th}$ regime AR lag order; $\phi_{0}^{(j)}$ is intercept and $\{\phi_{i}^{(j)}|i=1,2,\ldots, p_j \}$
are AR terms coefficients for $j^{th}$ regime; $I(.)$\footnote{$I(c)=1$ if $c$ is true, else 
	$I(c)=0$} is indicator function and $r=(r_{1}, \ldots, r_{k-1})$ 
satisfying $r_{1}< \ldots <r_{k-1}$ are the threshold values; $r_{0}=-\infty$ and 
$r_{k}=\infty$; $z_{t-d}$ is the threshold variable with positive integer delay parameter $d$; for each $j$, 
$\{\varepsilon_{t}^{(j)}\}$ is a sequence of martingale differences satisfying $\ex( \varepsilon_{t}^{(j)}|F_{t-1}) = 0$, 
${sup}_{t}\ex( |\varepsilon_{t}^{(j)}|^{\delta}|F_{t-1}) <\infty$ $almost$ $  surely$ for some $\delta>2$, where $F_{t-1}$ is the $\sigma$-field generated by $\{\varepsilon_{t-i}^{(j)}|i=1,2\ldots;j=1, \ldots, k\}$. 

For example, if at time $t$, $z_{t-d} = a \in (r_{j-1}, r_{j})$ then the active regime at that time is characterized by
\[
y_t = \phi_{0}^{(j)}+ \sum_{i=1}^{p_j}\phi_{i}^{(j)}y_{t-i} \ + \ \varepsilon_{t}^{(j)}.
\]

The model represented in Equation \ref{eq1} becomes a self-exciting threshold autoregressive (SETAR) model if the threshold variable is replaced by the time series variable itself, i.e., $z_{t-d}$ $=$ $y_{t-d}$  where the regime of the time series $y_{t}$ is now determined by its own past value $y_{t-d}$. 

Similar to the TAR model, a general SETAR($k$) model where $k$ denotes the number of regimes can be represented as
\begin{equation}\label{eq1_3}
	y_{t}  =\sum_{j=1}^{k} \left [ (\phi_{0}^{(j)} + \sum_{i=1}^{p_j}\phi_{i}^{(j)}y_{t-i} \ + \ \varepsilon_{t}^{(j)}) \ I(r_{j-1}<y_{t-d}\leq r_{j}) \right].
\end{equation}

The unknown parameters for the model in Equation \ref{eq1} are
\[
\Omega= \left [ (\phi_{0}^{1},\phi_{i:p_1}^{1}), \ldots,(\phi_{0}^{k},\phi_{i:p_k}^{k}), 
r=(r_{1}, \ldots, r_{k-1}),d \right ],
\] 
and can be estimated by a least-squares (LS) estimation under the assumption that 
$\varepsilon_{t}^{(j)}$ is i.i.d.$N(0,\sigma^2)$. The minimization of the sum of 
squared residuals yields the LS estimators: 
\begin{equation}\label{eq1_2}
	\widehat{\Omega}={\mbox{arg min}}_{\Omega} \sum_{t=1}^{T} \left [ y_{t}-\sum_{j=1}^{k}\left [(\phi_{0}^{(j)}+ \sum_{i=1}^{p_j}\phi_{i}^{(j)}y_{t-i})I(r_{j-1}<z_{t-d}\leq r_{j}) \right ] \right ]^2
\end{equation}
The minimization problem of Equation \ref{eq1_2} can be solved by a grid search over all combinations of possible values of parameters $\{j,p_{j},r,d|j=1, \ldots, k\}$. Thus, the search method requires a number of approximately $T^{k-1}\times d\times p_1\times \ldots \times p_k$ arranged auto-regressions. Alternatively, one can estimate the unknown thresholds and parameters via a model selection perspective by searching the minimum of a specified information criterion (e.g. AIC, BIC or HQIC). Fortunately, Gonzalo and Pitarakis \citep{gonzalo2002estimation} proposed a sequential model selection approach under an unknown number of thresholds for estimating all threshold parameters one at a time, which reduces  the computational cost significantly. The procedure starts with deciding between a linear and a one threshold (i.e. two regime) AR specification. If the existence of a threshold cannot be rejected then the sample can be arranged into two subsamples by the threshold. To search for the existence of the another threshold, the same procedure is repeated on both sub-samples that were conditionally created on the threshold in the first step. The iterations stop when the model selection procedure cannot verify the presence of an additional threshold. In this case, the required number of arranged auto-regressions to be estimated approximately decreased to $(T\times d\times p_1\times \ldots \times p_k + {(k-1)}\times T)$.

The statistical properties of the TAR model and more detailed information can be seen in \citep{chan1993consistency}, \citep{hansen1997inference} and \citep{amendola2009statistical}. In line with the potential of the TAR specification to distinguish multiple time series, we made use of the TAR specification for observing nonlinear associations, and the AR specification for observing linear associations in the multiple time series clustering task.

\subsection{Clustering of Time Series}
\label{sec:3.2}
The clustering of multiple univariate time series based on measured distances over raw data via common clustering methods, such as K-means and Fuzzy-C means, would be inappropriate since the raw time series data cannot exhibit most of its time-dependent statistical properties and structure without any statistical analyzing tool. Therefore, a proper way of summarizing the time-dependent data should be investigated to allow for comparing underlying structures.

In this study, the set of time series that needed to be clustered were represented by feature vectors. Feature vectors were specifically designated to cover the time-dependent information of the time series and the entries of each vector contained comparable model based outputs and time-dependent statistics that could be used to distinguish each time series. Nevertheless, real time series data may have a complicated generating mechanism (i.e. source) and features, and for this reason, we needed competent and flexible tools to capture and summarize them. In this respect, one of the concerns of the study was grouping time series with respect to the similarity between their true DGMs. The approximation to the DGM of each series was investigated using TAR models, which are known for their ability to represent nonlinear features in time series, such as abrupt changes and regime shifting behavior.

Forming a  feature vector from a univariate time series, $Y_s$, can be illustrated as; 
~~
\[ Y_s=\left[ \begin{array}{c}
y_1  \\
y_2  \\
\vdots\\
\vdots\\
\vdots\\
y_T \end{array} \right]_{T\times 1}====\Rightarrow  ~~~
f_s=\left[ \begin{array}{c}
\widehat{\phi}_{1\times {p_1}}^{(1)}  \\
\hdashline
\widehat{\phi}_{1\times {p_2}}^{(2)} \\
\hdashline
\vdots\\
\hdashline
\widehat{\phi}_{1\times {p_k}}^{(k)} \\
\hdashline
acf_{{res},{1\times l}}\\
\hdashline
pacf_{{res},{1\times l}}\\
\hdashline
ccf_{{res\times res^2},{1\times l}}\\
\hdashline
acf_{1\times l}\\
\hdashline
pacf_{1\times l}\\
\hdashline
\widehat{\phi}_{1\times {p_{ar}}}^{(ar)} \\
\end{array} \right]_{(p+5l)\times 1},
\]

where $p=p_1+ \ldots +p_k+p_{ar}$, and $l$ is the lag order that acf, pacf $\&$ $ccf$ are  calculated up to. To compare time series via TAR models, estimates of coefficients for each time series were included in feature vectors as, $\widehat{\phi}_{1\times {p_1}}^{(1)}$ $\ldots$ $\widehat{\phi}_{1\times {p_k}}^{(k)}$. The remaining serial correlation structure of residuals was observed by  $acf, pacf$ calculations and statistically significant correlations were added to the feature vector. The potential heteroscedasticity in the residuals was aimed to be evaluated by the  $ccf$ of residuals versus squared residuals and significant correlations were maintained at feature vectors. Finally, significant AR coefficients, $\widehat{\phi}_{1\times {p_{ar}}}^{(ar)}$ and the autocorrelation structure of the stationarized time series, $acf_{1\times l}$ $\&$ $pacf_{1\times l}$ that presents the linear associations were added to the feature vector. Thus, a time series with $T$ number of observations was represented as a point in the $p+5l$ dimensional space. Then, the spectral clustering approaches that are known for their superiority in graph partitioning would become available for the purpose of time series clustering.

In this study, we used the normalized spectral clustering in accordance with Ng et al.\citep{ng2002spectral}. The main idea of the procedure is using the eigenvectors of an affinity matrix that are derived from the data. The algorithm is summarized in Figure \ref{Spectral_clustering} (for the more theoretical aspects of spectral clustering see \citep{ng2002spectral} and \citep{von2007tutorial}):
\begin{figure*}
	{\setstretch{1.0}

		\begin{framed} 
			Given set of points $N=\{f_1,f_2,f_3,\ldots,f_n \}$ in $\mathbb{R}^v$ where  $v = p+5l$ in this study, can be clustered into $k$ subsets: 
			
			\begin{enumerate} \small{
					
					\item Constructing the $n\times n$ Affinity matrix  $A$ which is defined by \[A_{ij}=\frac{exp(-\parallel f_i-f_j \parallel^2)}{(2\sigma^2 )}\] if $i\neq j$, and  $A_{ii}=0$
					
					\item Define matrix $D$ to be diagonal matrix whose $(i,i)$ element is the sum of  $A$\vtick s  $i^{th}$ row, and compute the normalized $L$aplacian matrix  $L=D^{-1/2}  A D^{-1/2}$
					
					\item Construct the eigenvector matrix  $X_{n\times k}$ from $k$ largest eigenvectors of  $L$ that are stacked in columns of $X$.
					
					\item Compute the matrix $U$ from re-normalized rows of the matrix $X$.  
					
					\item Cluster each row of $U$ as a point in $\mathbb{R}^k$ into $k$ clusters by any clustering algorithm.
					
					\item Assign the original point $f_i$ to cluster number $j$ iff row $i$ of the re-normalized eigenvector matrix $U$ was assigned to cluster number $j$.
				}
			\end{enumerate}
		\end{framed}
	}
	\caption{Spectral clustering}
	\label{Spectral_clustering}
\end{figure*}
\subsection{Proposed time series clustering approach}
\label{sec:3.3}

Describing the temporal dynamics of a time series process can be achieved by utilizing TAR modeling and, hence, the logic of the use of TAR model outputs in time series clustering is somewhat akin to imposing a similar effect to that of centrifuge machinery (i.e. a separator tool for different substances with respect to their densities) throughout the observed multiple time series. Rather than finding an exact or "true" model for each process, the main motivation was forming a plausible scale that allows us to distinguish time series according to their associations with different phases of the TAR model. In addition, linear AR model outputs and autocorrelation information were also considered during the formation of the feature space. 

The proposed clustering approach aimed to produce a reasonable partition of time series by their feature vectors, which were designated to contain associations of time series with linear and non-linear properties. As we state in Section \ref{sec:2}, the temporal properties of time series were refined in the feature space, which has more explicit characteristics (i.e. spectra) than raw time series data.

Steps of the proposed approach given in Figure \ref{clustering_steps}:

\begin{figure*}
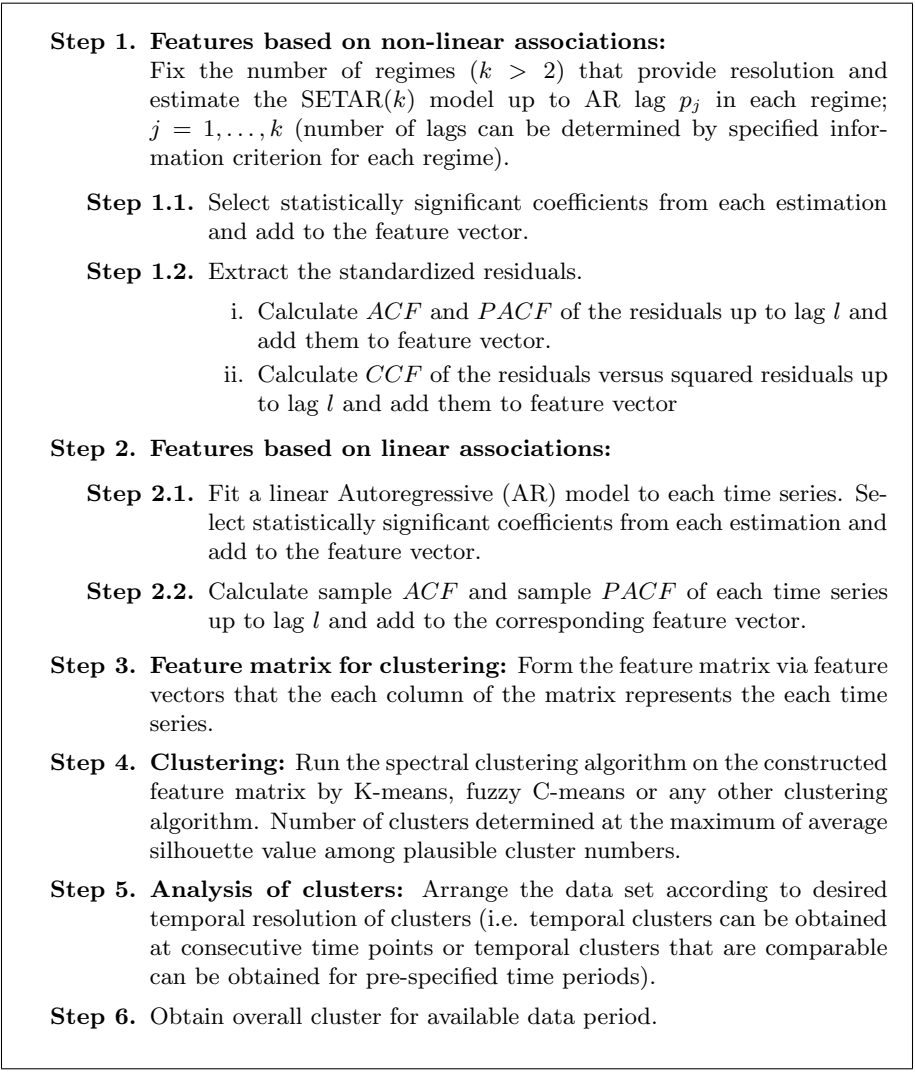

	{\setstretch{1.0}

		\begin{framed}\small{
				\begin{enumerate}[label=\bfseries Step \arabic*.,leftmargin=*,labelindent=1em]
					\item \textbf{Features based on non-linear associations:} \\
					Fix the number of regimes ($k>2$) that provide resolution and estimate the SETAR($k$) model up to AR lag $p_j$ in each regime; $j=1,\ldots,k$  (number of lags can be determined by specified information criterion for each regime).
					\begin{enumerate}[label*=\bfseries \arabic*.]
						\item 	Select statistically significant coefficients from each estimation and add to the feature vector.
						\item 	Extract the standardized residuals.
						\begin{enumerate}
							\item Calculate $ACF$ and $PACF$ of the residuals up to lag $l$ and add them to feature vector.
							\item Calculate $CCF$ of the residuals versus squared residuals up to lag $l$ and add them to feature vector
						\end{enumerate}
					\end{enumerate}
					\item \textbf{Features based on linear associations:} 
					\begin{enumerate}[label*=\bfseries \arabic*.]
						\item 	Fit a linear Autoregressive (AR) model to each time series. Select statistically significant coefficients from each estimation and add to the feature vector.
						\item Calculate sample $ACF$ and sample $PACF$ of each time series up to lag $l$ and add to the corresponding feature vector.
					\end{enumerate}
					\item \textbf{Feature matrix for clustering:} Form the feature matrix via feature vectors that the each column of the matrix represents the each time series. 
					\item	\textbf{Clustering:} Run the spectral clustering algorithm on the constructed feature matrix by K-means, fuzzy C-means or any other clustering algorithm. Number of clusters determined at the maximum of average silhouette value among plausible cluster numbers.
					\item	\textbf{Analysis of clusters:} Arrange the data set according to desired temporal resolution of clusters (i.e. temporal clusters can be obtained at consecutive time points or temporal clusters that are comparable can be obtained for pre-specified time periods). 
					\item	Obtain overall cluster for available data period. 
				\end{enumerate}
			}
		\end{framed}
	}
	\caption{Steps of the proposed time series clustering}
	\label{clustering_steps}
\end{figure*}
\section{Computational Study}
\label{sec:4}

To assess the performance and the ability of the proposed time series clustering approach, the simulation study and applications on real data sets were conducted and the details are given in Section \ref{sec:4.1} and \ref{sec:4.2}. 

The applicability of the proposed time series clustering approach was evaluated by considering various linear and non-linear time series models (i.e. DGMs) in the simulation scenario. To determine the optimum number of clusters, the average silhouette values were used.

With the support of the results from the simulation study, the proposed time series clustering approach was applied on several commodity prices to get price groups that share similar dynamics. In addition to this, it was possible to check whether the price groups varied across time. Commodity price clusters and their time-dependent behaviors were also evaluated, see Section \ref{sec:4.2}.

\begin{table}[]
	\caption{Simulated models (i.e DGM).}
	\label{tab_sim}
	\setlength{\tabcolsep}{4.75pt}
	\def\arraystretch{1}
	\begin{tabular}{lrr}
		\specialrule{.25em}{.25em}{.25em} 
		&&\\
		ser01 & \(Y_t=0.80Y_{t-12}+\varepsilon_t+0.70\varepsilon_{t-12} \) & $\varepsilon_t$ $\sim N(0,1)$  \\
		&&\\
		ser02 & \(Y_t=-0.70Y_{t-24}+\varepsilon_t+0.80\varepsilon_{t-6} \) & $\varepsilon_t$ $\sim N(0,1)$ \\
		&&\\
		ser03 & \(Y_t=0.80Y_{t-1}-0.40Y_{t-2}+0.15Y_{t-3}+\varepsilon_t-0.20\varepsilon_{t-1}+0.25\varepsilon_{t-2} \) & $\varepsilon_t$ $\sim N(0,1)$ \\
		&&\\
		ser04 & \(Y_t=0.90Y_{t-1}-0.80Y_{t-2}+0.55Y_{t-3}+\varepsilon_t+0.80\varepsilon_{t-1}+0.50\varepsilon_{t-2} \) & $\varepsilon_t$ $\sim N(0,1)$ \\
		&&\\
		ser05 & \(Y_t=1.10Y_{t-1}-0.60Y_{t-2}-0.20Y_{t-3}+\varepsilon_t+0.30\varepsilon_{t-1}-0.70\varepsilon_{t-2} \) & $\varepsilon_t$ $\sim N(0,1)$ \\
		&&\\
		ser06 & \(Y_t=2.55Y_{t-1}-2.30 Y_{t-2}+0.75Y_{t-3}+\varepsilon_t+0.80\varepsilon_{t-1}+0.50\varepsilon_{t-2} \) & $\varepsilon_t$ $\sim N(0,1)$ \\
		&&\\
		ser07 & 
		\begin{tabular}{lcrr}
			& \multirow{3}{*}{$\Bigg\{$} &\({2 - 0.40 Y_{t-1}- 0.10 Y_{t-2}+\varepsilon_{t}},~~~~~~~\) & ${Y_{t-1}<-1}$\\ 
			$Y_t=$&& \(-0.05 + 0.20 Y_{t-1}+ 0.70 Y_{t-2}+ \varepsilon_{t},~~~~~~~\) & $-1 <$ ${Y_{t-1}\leq 1}$ \\
			&& \(0.05 - 0.45 Y_{t-1}+ 0.15 Y_{t-2}+\varepsilon_{t},~~~~~~~\) & ${Y_{t-1}\geq 1}$ 
		\end{tabular} 
		& $\varepsilon_t$ $\sim N(0,1)$ \\
		&&\\
		ser08 & 
		\begin{tabular}{lcrr}
			& \multirow{3}{*}{$\Bigg\{$} &\({-0.50 + 0.40 Y_{t-1}- 0.10 Y_{t-2}+\varepsilon_{t}},~~~~~~~~~\) & ${Y_{t-1}<0}$\\ 
			$Y_t=$&& \(0.05 + 0.20 Y_{t-1}+ 0.80 Y_{t-2}+ \varepsilon_{t},~~~~~~~~~\) & $0 <$ ${Y_{t-1}\leq 4}$ \\
			&& \(0.05 - 0.45 Y_{t-1}+ 0.15 Y_{t-2}+\varepsilon_{t},~~~~~~~~~\) & ${Y_{t-1}\geq 4}$ 
		\end{tabular} 
		& $\varepsilon_t$ $\sim N(0,1)$ \\
		&&\\
		ser09 & 
		\begin{tabular}{lcrr}
			& \multirow{3}{*}{$\Bigg\{$}	&\({-0.15 + 0.74 Y_{t-1}- 0.15 Y_{t-2}+\varepsilon_{t}},~~\) & ${Y_{t-1}<-1.2}$\\ 
			$Y_t=$&& \(1.90 + 0.20 Y_{t-1}-1.30 Y_{t-2}+ \varepsilon_{t},~~\) & $-1.2 <$ ${Y_{t-1}\leq 1.2}$ \\
			&& \(1.00 + 0.50 Y_{t-1} - 1.15 Y_{t-2}+\varepsilon_{t},~~\) & ${Y_{t-1}\geq 1.2}$ 
		\end{tabular} 
		& $\varepsilon_t$ $\sim N(0,1)$ \\
		&&\\
		ser10 & 
		\begin{tabular}{lcrr}
			& \multirow{3}{*}{$\Bigg\{$}&\({3 + 0.50 Y_{t-1}- 0.80 Y_{t-2}+0.40 Y_{t-3}+\varepsilon_{t}}\), & ${Y_{t-1}<3}$\\ 
			$Y_t=$&& \(6.00 + 0.90 Y_{t-1}+ \varepsilon_{t},\) & $3 <$ ${Y_{t-1}\leq 9}$ \\
			&& \(4.00 + 0.70 Y_{t-1} - 0.80 Y_{t-2}+\varepsilon_{t},\) & ${Y_{t-1}\geq 9}$ 
		\end{tabular} 
		& $\varepsilon_t$ $\sim N(0,1)$ \\
		&&\\
		\specialrule{.25em}{.25em}{.25em} 
	\end{tabular}
	
\end{table}

\subsection{Simulation}
\label{sec:4.1}

\begin{figure*}
	\includegraphics[width=1 \textwidth]{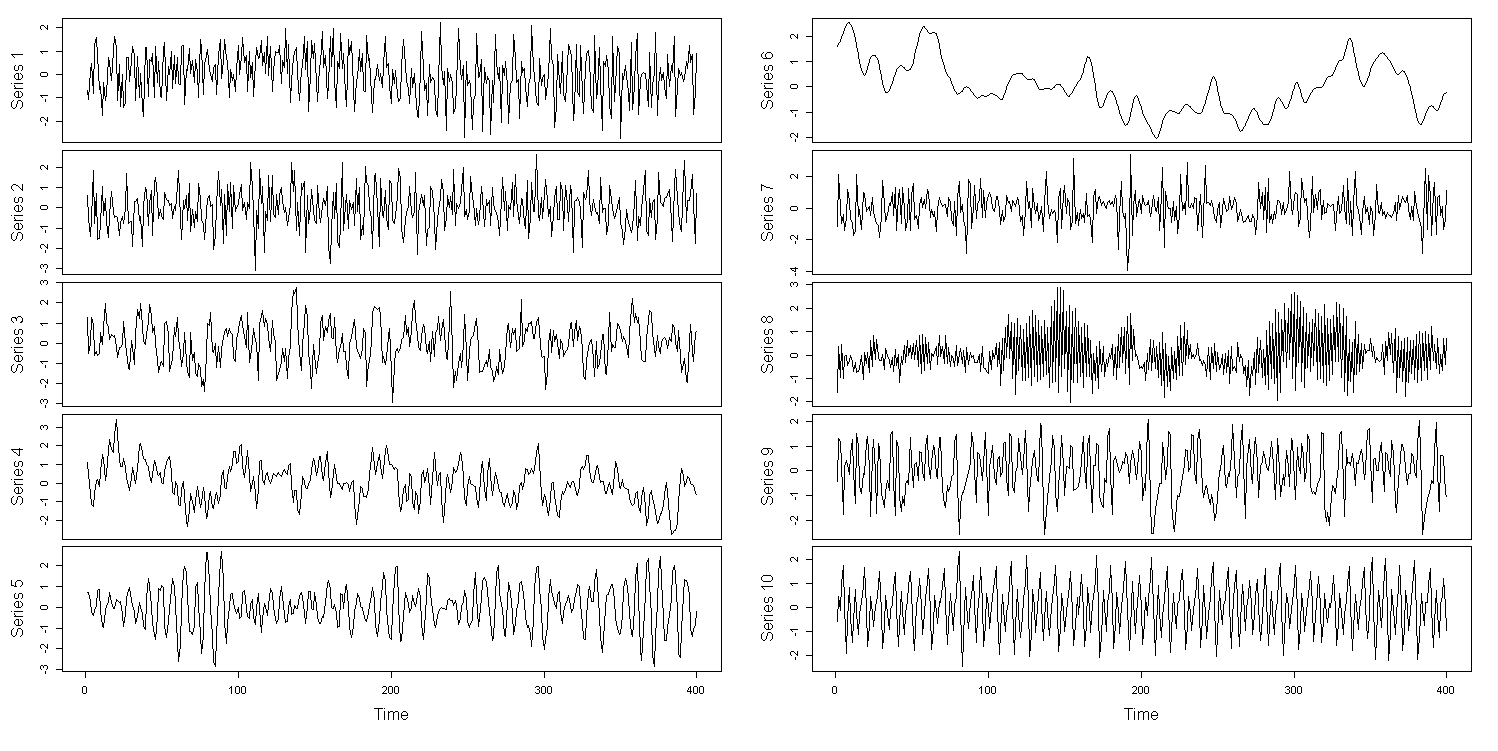}
	\caption{Simulation sample from 10 different DGMs.}
	\label{fig_sts}       
\end{figure*}

The simulation study was conducted to evaluate the effectiveness of the proposed time series clustering approach. For the simulation study, consideration was given to 10 different time series models (i.e. DGMs), given in Table \ref{tab_sim}, consisting of 400 observations each. Representative patterns of the DGMs are shown in Figure \ref{fig_sts}. Ten different samples were generated from each model and a final dataset of size $10\times 10\times 400$ was clustered with the proposed clustering approach. To explore the sampling variability of clustering results, this scenario was replicated 30 times.

The DGMs used in the study were selected to cover different types of characteristics of time series, such as stationarity, non-stationarity, seasonality and threshold non-linearity or regime switching. In order to assess the sensitivity of the proposed approach over similar DGMs, some of them were selected based upon their similarity in terms of dynamics/structure, such as lag orders and coefficients. The first 6 models in Table \ref{tab_sim} are from the family of linear processes, more specifically: the 1\textsuperscript{st} and 2\textsuperscript{nd} ones have mainly seasonal characteristics; the 3\textsuperscript{rd}, 4\textsuperscript{th} and 5\textsuperscript{th} ones share the same orders but with slightly different magnitudes at each order; and the 6\textsuperscript{th} model is the first order integrated non-stationary process. The last four models in Table \ref{tab_sim} are three regime threshold non-linear processes with different magnitudes and threshold values in each regime.

In order to determine the appropriate number of clusters, the silhouette method developed by Rousseeuw \citep{rousseeuw1987silhouettes} was considered. For each object $i$ (time series variable or corresponding feature vector in our case), one can find a certain value  $s(i)$ that is called the silhouette value and it can be calculated by,
\[s(i)=\frac{b(i)-a(i)}{max\{a(i),b(i)\}},\]
\[-1 \leq s(i) \leq 1 ,\]

where $a(i)$ is the average dissimilarity of object $i$ to all other objects within the same cluster, $b(i)$ is the minimum of average dissimilarities of object $i$ to all other objects in other clusters. The silhouette value, $s(i)$, attains its maximum value when the best clustering is observed for the object $i$. The maximum possible value of  $s(i)$ is 1 and this occurs when the \textit{within dissimilarity} $a(i)$ is much smaller (close to $0$) than the smallest \textit{between dissimilarity}, $b(i)$.

For a specific number of clusters, $c$,  the overall quality of clustering can be evaluated by assessing the average value of $s(i)$, \(\bar S_c=\frac{1}{n}\sum_{i}^{n}s(i)\) where $n$ is the total number of objects (time series or corresponding feature vector) to be clustered. Thus, the number of optimum clusters can be determined by finding the maximum of the average silhouette value, $\bar S$,  amongst the possible numbers of clusters.

\begin{figure*}
	\includegraphics[width=1 \textwidth]{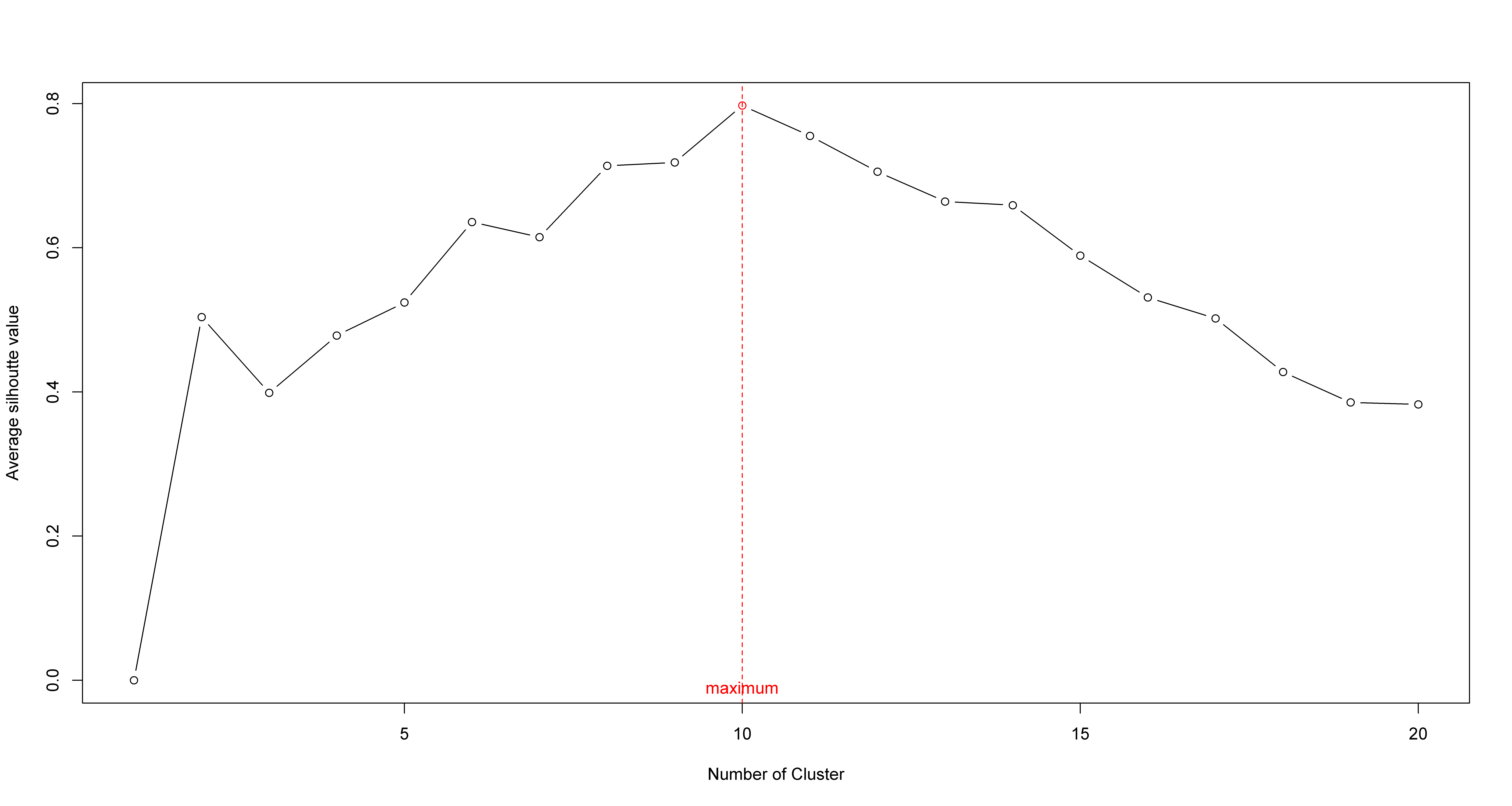}
	\caption{Average silhouette value per cluster number for a replicate.}
	\label{fig_asv}       
\end{figure*}

\begin{figure*}
	\includegraphics[width=1 \textwidth]{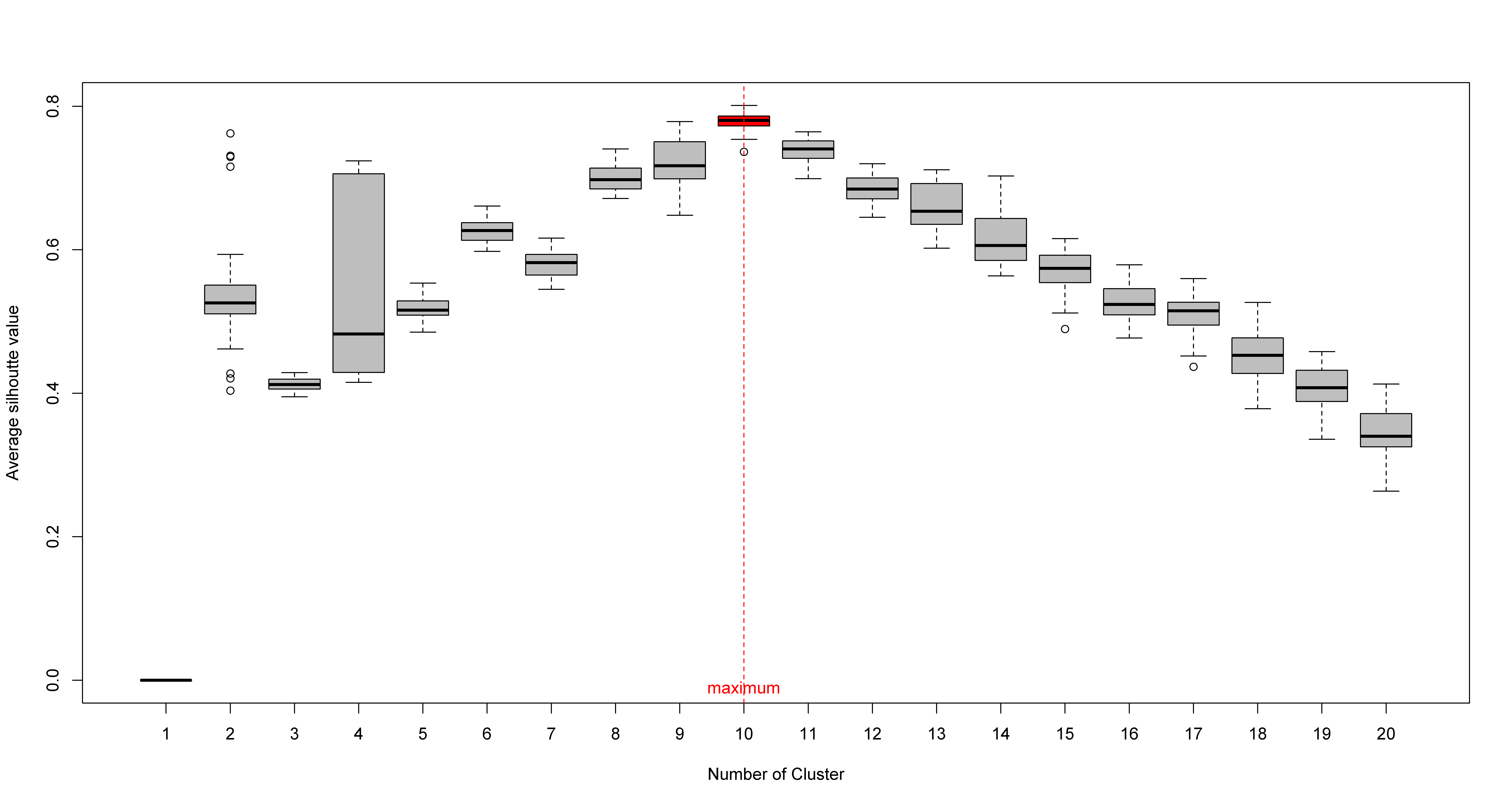}
	\caption{Distribution of average silhouette values per cluster number for 30 replicates.}
	\label{fig_asvps}       
\end{figure*}

Figure \ref{fig_asv} presents the $\bar S$ values for one replication of the simulation scenario and it indicates the optimum number of clusters being selected at the maximum $\bar S$ value, which is the actual number of DGM group. In order to explore the sampling variability on $\bar S$ values, 30 independent replicates of the simulation scenario were considered and the resultant box-plots per cluster number are presented in Figure \ref{fig_asvps}. In this figure, the maximum $\bar S$ values with the smallest variance were obtained at cluster number 10, which is the true number of DGM group. Besides its noticeable distribution and large variance, the box-plots for cluster numbers 2 and 4 implied that the $\bar S$ values could attain such a value near the maximum that we could refer to them as local maximums. This result for the simulation study logically corresponded to the general categories of the selected models. That is to say, the considered models (DGMs) in the simulation study could be classified as 2 main groups, such as linear vs non-linear models, which explained the outliers of box-plot at cluster number 2 or they could also be classified into 4 categories, such as stationary, non-stationary, seasonal and non-linear models, which explained the skewness of box-plot at cluster number 4.  

\begin{figure*}
	\includegraphics[width=1 \textwidth]{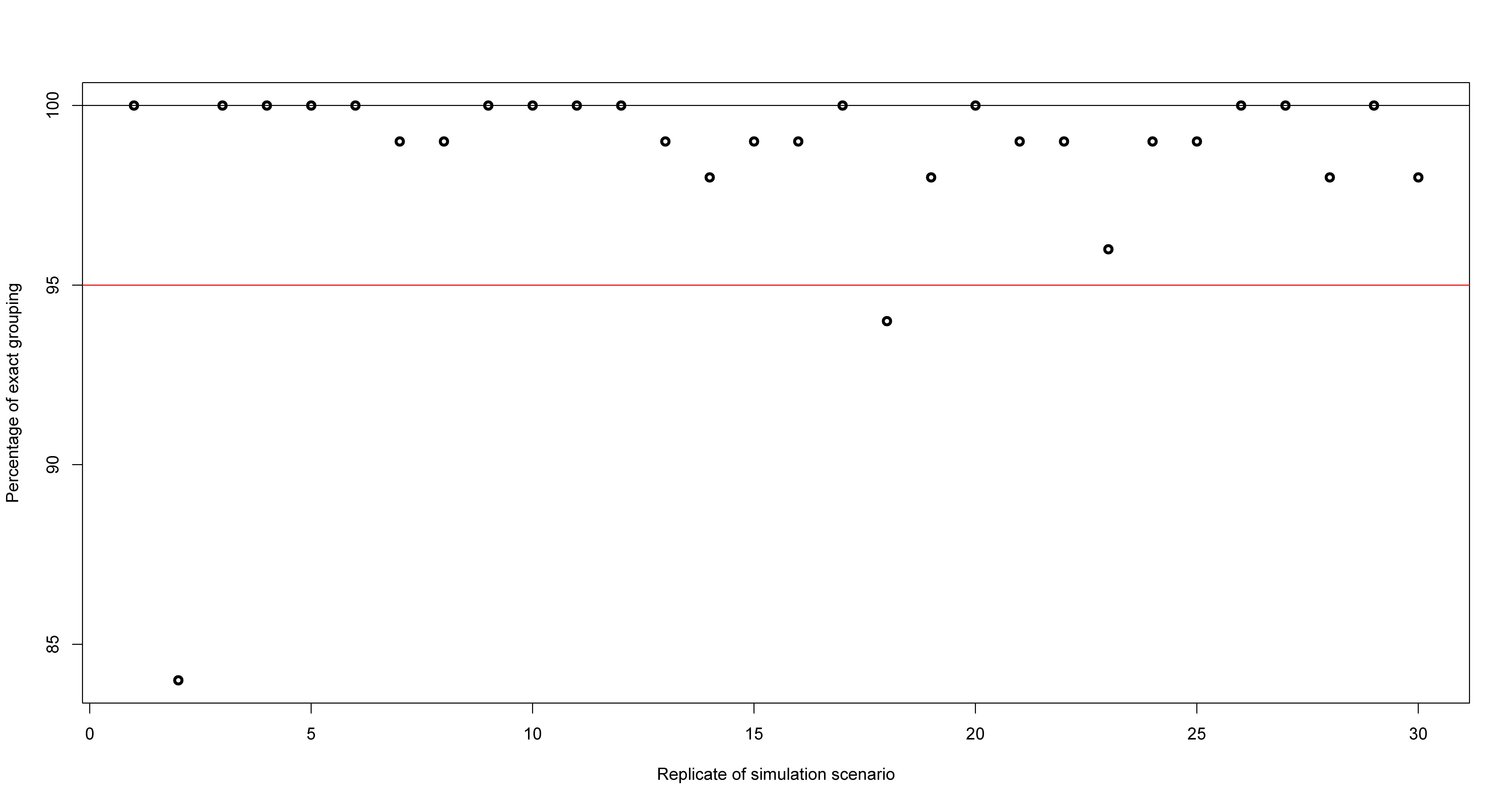}
	\caption{Exact grouping percentages over 30 replicates.}
	\label{fig_gp}       
\end{figure*}

The simulation study revealed that the true number of clusters could be determined via the proposed approach by finding the maximum average silhouette value over possible cluster numbers. With the optimum number of clusters set at 10, the correct assignments of the simulated time series to their true DGM group for 30 replicates of the simulation scenario are given in Figure \ref{fig_gp}. According to the overall result, 98.5\% of the simulated time series clustered into their true DGM group, given in Table \ref{tab_sim}.

\subsection{Application to commodity prices}
\label{sec:4.2}
\begin{figure}
	\begin{center}
		\includegraphics[width=1\textwidth]{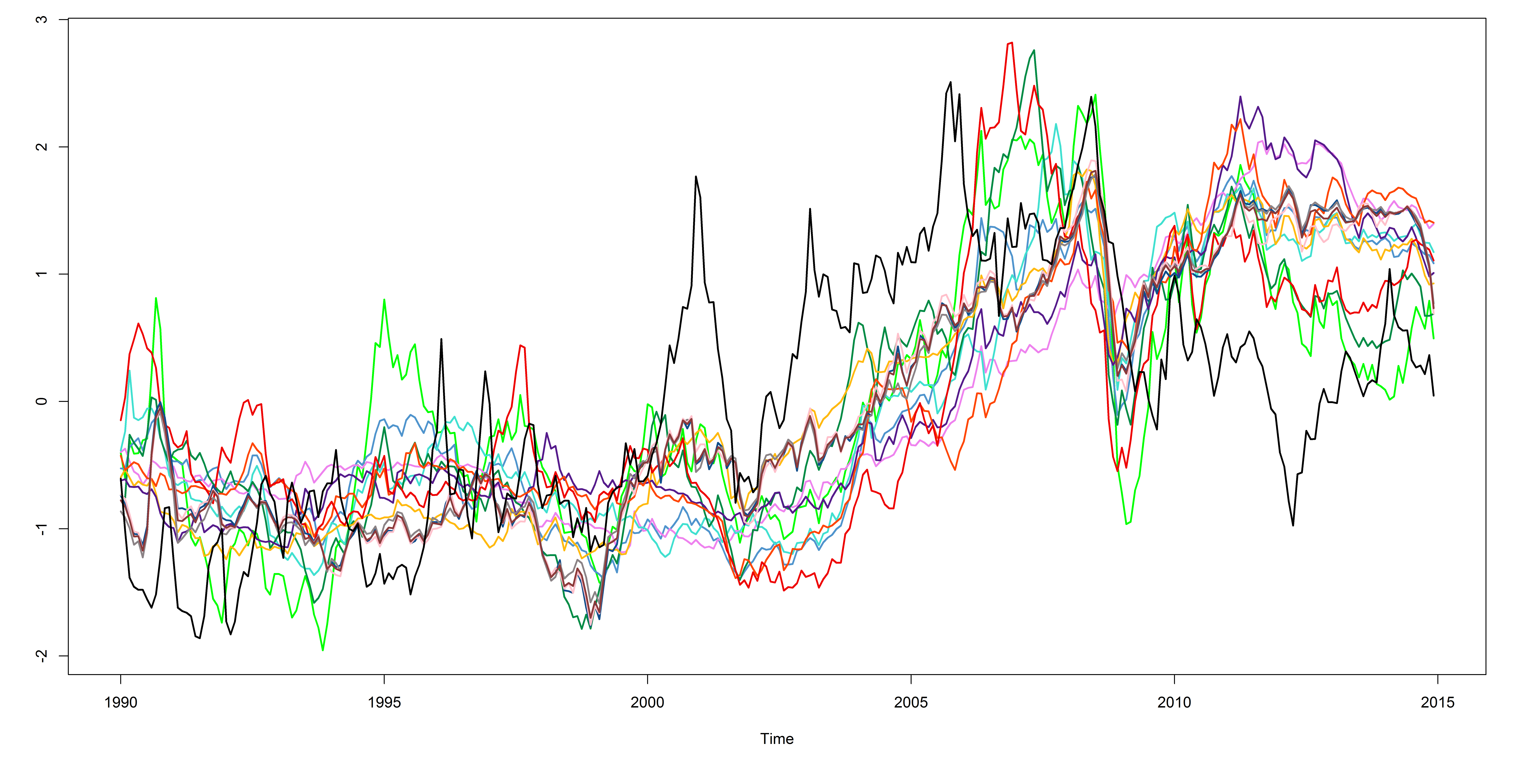}
		\includegraphics[width=0.40 \textwidth]{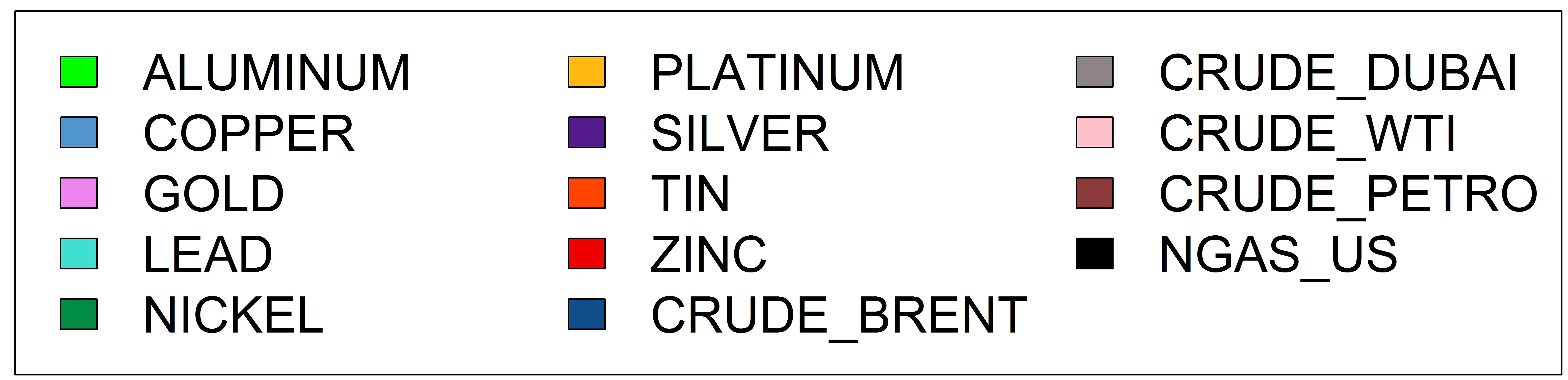}
	\end{center}
	\caption{Plots of logged and scaled commodity prices.}
	\label{fig_ts}       
\end{figure}

The data used in the study contained the monthly averages of 14 commodity prices between January 1990 and December 2014. The data were obtained from the publicly available database of the World Bank for Commodity Price Dataset - Global Economic Monitor (GEM) Commodities. Table \ref{tab_ts} shows the prices used in the study and Figure \ref{fig_ts} depicts the logged and scaled traces of each commodity price. 

\begin{table}
	\caption{Commodity prices used in the study.}
	\label{tab_ts} 
	\centering
	\begin{tabular}{ll}
		\specialrule{.25em}{.25em}{.25em}
		\noalign{\smallskip}
		\bf{Metals} & \bf{Energy}  \\
		\noalign{\smallskip}\hline\noalign{\smallskip}
		Aluminum & Crude Oil Brent\\ 
		Copper & Crude Oil WTI \\
		Lead & Crude Oil Dubai \\
		Nickel & Crude Oil Avg \\
		Tin & Natural Gas US\\
		Zinc &  \\
		Gold & \\
		Platinum & \\
		Silver &\\
		\noalign{\smallskip}
		\specialrule{.25em}{.25em}{.25em} 
	\end{tabular}
\end{table}

Here, we considered samples from two general categories of the commodities, namely metal and energy related commodities. As we state in Section \ref{sec:3.2} and \ref{sec:3.3}, in determining the price groups, the proposed clustering approach would not impose any qualitative information related to price categories. We aimed to find clusters that contained similar commodities in terms of their price behaviors. In this context, the comparing and clustering of price behaviors were independent from their categorical classification. However, the viability of the qualitative categorization could be investigated by clustering over the intra (i.e. metal prices) and inter (i.e. metal \& energy prices) categories.

Moreover, to illustrate the time varying price clusters, the proposed clustering approach was implemented on three different time periods. The first period, from January 1990 to December 2007, was chosen to be outside the instant effect of the global financial crisis that occurred in 2008. The second period, from January 2010 to December 2014, was chosen to see whether the behavior of price groups changed after the global financial crisis. Finally, the third period, spanning the whole period and including the effect of the financial crisis, was used to compare cluster results with the clusters obtained in the first period.

One of the important steps of the proposed approach was the feature extraction based on non-linear associations. This task was accomplished by fitting a  SETAR($k$) model to each commodity price. The choice of the number of regimes for economic time series can be decided by the framework of the empirical study, and the number of regimes were generally considered as two \citep[see, for example,][]{charles2015new, ferraresi2015fiscal, hamilton2016macroeconomic}. It should be noted that, for the sake of clarity, the number of regimes, $k$, was fixed to three for both the simulated and real data sets. From a statistical point of view, the SETAR concept for modeling any time series can be utilized after ensuring the existence of a threshold non-linearity. However, the aim of using the SETAR specification in the proposed approach was not for modeling and forecasting, but rather for measuring the associations of time series with different phases of the SETAR model by the coefficients and correlation structure of the residuals that were obtained from the estimation of SETAR($k$).  Therefore, proving the existence of a threshold non-linearity (i.e. $k>1$) was not strictly necessary in this case. As we observed in the simulation study, using the SETAR model as a seperator tool in the proposed approach performed well even for similar linear causal time series models. Nevertheless, although it was not necessary with regard to implementing the proposed clustering approach, we evaluated the regime shifting behavior or threshold non-linearity of the commodities using Hansen's threshold non-linearity test that was developed by Hansen \citep{hansen1999testing}. Table \ref{tab_th} shows the $p-$values of the test results. According to the test results, the threshold non-linearity of commodity prices and regime shifting behavior could not be rejected at 0.05 significance level except for ZINC and CRUDE-WTI.

\begin{table}[]
	\caption{Hansen\vtick s threshold non-linearity test results ($P-$Values).}
	\label{tab_th}
	\setlength{\tabcolsep}{6pt}
	\def\arraystretch{1.05}
	\begin{tabular}{lccc}
		\specialrule{.25em}{.25em}{.25em} 
		{}&$TAR(1) vs TAR(2)$&$TAR(1) vs TAR(3)$&$TAR(2) vs TAR(3)$\\
		\hline
		$ALUMINUM$&$<0.01$&$<0.01$&$\phantom{<}0.12$\\
		$COPPER$&$\phantom{<}0.05$&$\phantom{<}0.05$&$\phantom{<}0.07$\\
		$GOLD$&$<0.01$&$<0.01$&$<0.01$\\
		$LEAD$&$\phantom{<}0.02$&$\phantom{<}0.02$&$\phantom{<}0.06$\\
		$NICKEL$&$\phantom{<}0.05$&$\phantom{<}0.27$&$\phantom{<}0.91$\\
		$PLATINUM$&$\phantom{<}0.03$&$\phantom{<}0.09$&$\phantom{<}0.55$\\
		$SILVER$&$<0.01$&$<0.01$&$\phantom{<}0.01$\\
		$TIN$&$\phantom{<}0.05$&$\phantom{<}0.02$&$\phantom{<}0.10$\\
		$ZINC$&$\phantom{<}0.38$&$\phantom{<}0.34$&$\phantom{<}0.44$\\
		$CRUDE-BRENT$&$\phantom{<}0.03$&$\phantom{<}0.13$&$\phantom{<}0.92$\\
		$CRUDE-DUBAI$&$\phantom{<}0.01$&$\phantom{<}0.02$&$\phantom{<}0.10$\\
		$CRUDE-WTI$&$\phantom{<}0.10$&$\phantom{<}0.23$&$\phantom{<}0.58$\\
		$CRUDE-PETRO$&$\phantom{<}0.04$&$\phantom{<}0.07$&$\phantom{<}0.26$\\
		$NGAS-US$&$<0.01$&$\phantom{<}0.02$&$\phantom{<}0.28$\\
		\specialrule{.25em}{.25em}{.25em} 
	\end{tabular}
	
\end{table}

\begin{figure*}
	\includegraphics[width=1 \textwidth]{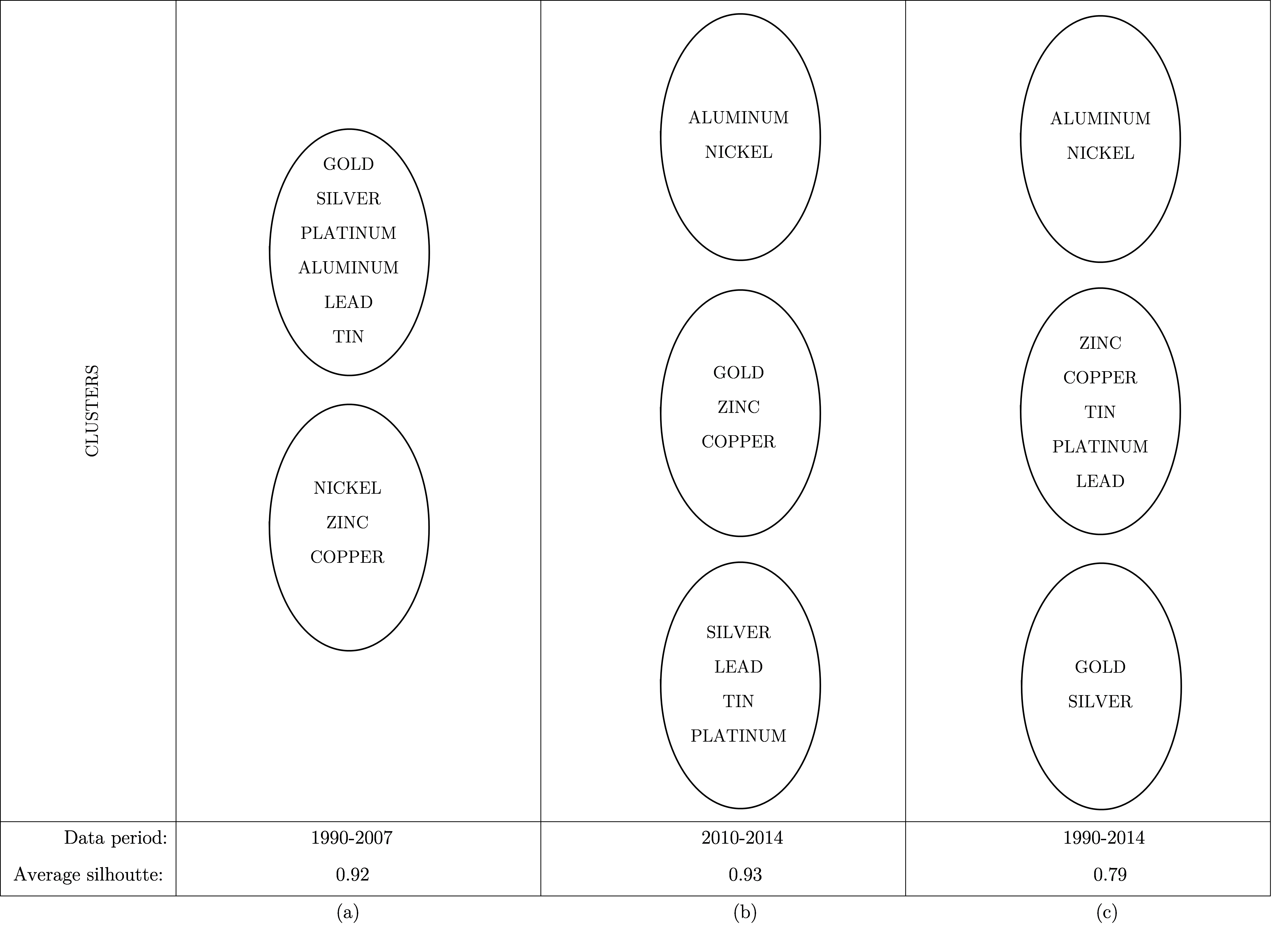}
	\caption{Temporal clusters obtained for metals only: (a) before crisis (b) after crisis (c) whole data span, including the effect of the crisis.}
	\label{fig_metals}       
\end{figure*}

\begin{figure*}
	\includegraphics[width=1 \textwidth]{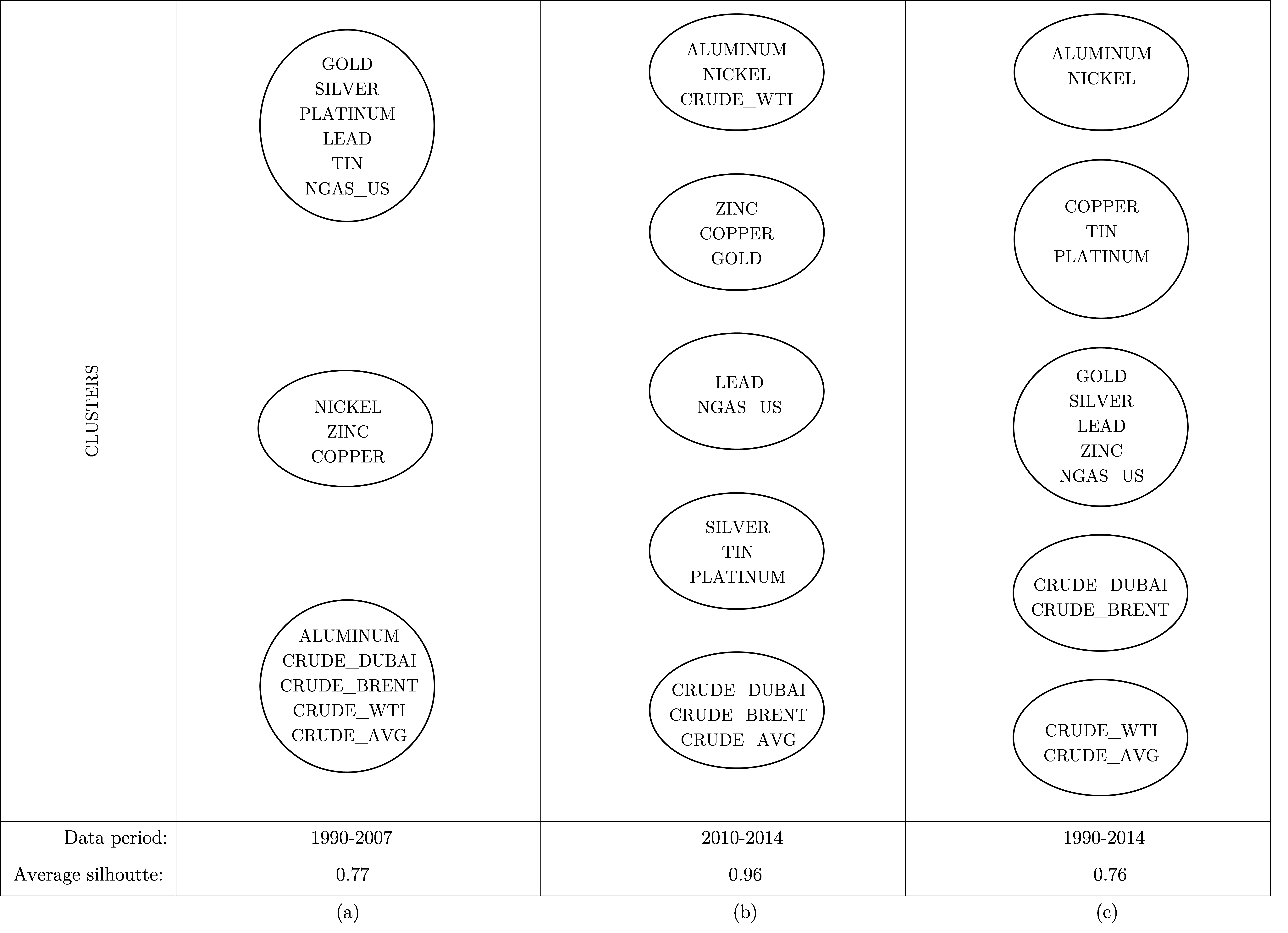}
	\caption{Temporal clusters obtained for all price series: (a) before crisis (b) after crisis (c) whole data span, including the effect of the crisis.}
	\label{fig_metals_energy}       
\end{figure*}

The results of metal prices and metal \& energy prices clustering are shown in Figure \ref{fig_metals}(a)-(c) and Figure \ref{fig_metals_energy}(a)-(c). Each cluster implied that the cluster members came from a similar source of data generating mechanisms, and the co-movement of commodities could be valid for each cluster members.

The general category of metals in the commodity market was qualitatively divided into two sub categories: base and precious metals. For the sake of clarity, aluminum, lead, tin, nickel, zinc and copper are base (i.e. industrial) metals, whereas gold, silver and platinum are precious metals. Figure  \ref{fig_metals}(a) shows the two groups obtained with maximum $\bar S$ at 0.92 for metal prices over the first period (i.e. before global financial crisis). Here, one of the groups consists of all the precious and three of the base metals (aluminum, lead and tin). Figure \ref{fig_metals}(b) shows the period after the global financial crisis: the number of groups increased to three and the group members changed. Three clusters were again obtained when we considered the overall data period, shown in Figure \ref{fig_metals}(c). It should be noted that both the number of clusters and the intra category (i.e metals) transitions were being observed across time. 

Extending the data set to include five energy related commodities caused slightly different cluster formations than in the previous configuration. In Figure \ref{fig_metals_energy}(a), (b) and (c) coincide with their counterparts in Figure \ref{fig_metals} and share some similarities. However, considering the dynamic properties of energy related commodity prices with the metal prices together affected the cluster numbers and members. For example, when we look at the first data period in Figure \ref{fig_metals_energy}, aluminum formed a cluster with crude oil prices, whereas in Figure \ref{fig_metals}(a) it was clustered with five other  metal prices. This implies that the dynamic characteristics of the aluminum price are more similar to the crude oil price than the other metal prices in the first period. Similarly, the lead prices formed a cluster with the natural gas prices in the second period, as shown in Figure \ref{fig_metals_energy}(b), which is different from when only metal prices were considered, as shown in Figure \ref{fig_metals}(b). In this respect, the results of the clustering were dependent on the selected commodities, which were chosen in line with the aims of the study. Although testing the hypothesis of the co-movement between commodity prices was not the primary aim, the scope of the study indicated limited similarities with the previous studies that focused on the co-movement of commodities, such as
\citep{lescar09, poncela2014common,matesanz2014co,fer15,ross15}. Since determining the price groups in this study relied on a comparison between their generating mechanisms, we could infer that the group members that shared similar dynamics tend to co-move over a certain period. In this context, our findings do not support the hypothesis of herding behavior of commodity prices, and we observe that the co-movement of the commodity prices can only be valid for time-dependently determined price groups (i.e. clusters).

\section{Concluding remarks}
Time series clustering and time varying clusters were investigated in this study. We proposed the clustering approach based on TAR and AR specifications that are able to collect the non-linear and linear associations of time series, respectively. To this end, the feature vectors were formed that were expected to represent the linear and non-linear temporal properties of time series. More importantly, feature vectors (or points) can be regenerated at each time index. In other words, time series are represented as points in the formed feature space in such a way that the dynamics of the points can be traceable across the time. Thus, the time varying groups of time series can be found by partitioning those points across time. Naturally, the points close to each other must be considered to be in the same group. Spectral clustering by Ng et al. \citep{ng2002spectral} is adapted to a time series clustering task because of its known merits with regard to the partitioning of a set of points in $\mathbb{R}^{n}$.

The  proposed clustering approach was evaluated according to a simulation study and applied on several commodity price series to obtain time varying clusters. It reveals that feature vectors based upon the time series models (e.g. TAR and AR models) can be used for the capturing and clustering of temporal dependencies of time series that originate from different sources. Determining the differences between multiple time series can be better facilitated if the models used in the clustering task are able to capture frequently encountered characteristics in time series, such as seasonality, asymmetry, regime shifting, time varying and chaotic behaviors. Then, the association of each time series with the considered time series models can be obtained and such strong associations can be emphasized by feature extraction. In fact, TAR models are appropriate for extending it  with more components such as ARCH/GARCH structure in the innovations. Ultimately, the proposed approach suggests the use of omni-directional time series models for time series clustering, which can provide outputs such as those that are used as separators for multiple time series. 

The obtained commodity price clusters are promising and, therefore, the proposed approach can be used to generate specific price group indexes such as RICI, TRCCI and S\&P GSCI. The results provided from clustering can serve as a basis for further multivariate statistical analyses such as VAR modeling and co-integration.

There are several possibilities for future work. The effectiveness of the proposed approach should be examined by considering the different kinds of DGMs. For example, using ARCH/GARCH effects with threshold models or hybrid models such as double threshold GARCH models can be considered. The number of variables used in the study was selected to give the longest coverage period and we used the monthly averages of commodity prices. The effectiveness of the proposed approach needs to be evaluated for high frequency time series such as daily and hourly data. Finally, expanding the feature space by constructing feature matrices for each time series and multivariate counterparts can also be considered.

\bibliography{referencesarx}

\end{document}